\documentclass[manuscript,screen,nonacm]{acmart}
\AtBeginDocument{%
  \providecommand\BibTeX{{%
    \normalfont B\kern-0.5em{\scshape i\kern-0.25em b}\kern-0.8em\TeX}}}

\acmDOI{}

\acmConference[HRCI]{Workshop on Human-Robot Conversational Interaction}{13 March 2023}{Stockholm, Sweden}
\acmBooktitle{Workshop on Human-Robot Conversational Interaction, 13 March 2023, Stockholm, Sweden} 
\begin{document}

\title[Who's in Charge? Decision-Making in Conversational Robots]{Who's in Charge? Roles and Responsibilities of Decision-Making Components in Conversational Robots}

\author{Pierre Lison}
\email{plison@nr.no}

\affiliation{%
  \institution{University of Oslo}
  \streetaddress{Problemveien 7}
  \city{Oslo}
  \country{Norway}
  \postcode{0315}
}
\author{Casey Kennington}
\email{caseykennington@boisestate.edu}
\affiliation{%
  \institution{Boise State University}
  \streetaddress{1910 W University Dr}
  \city{Boise}
  \state{Idaho}
  \country{USA}
  \postcode{83725}
}

% \affiliation{%

% \affiliation{%

% \affiliation{%

% \affiliation{%

% \affiliation{%

% \affiliation{%

% \affiliation{%

\renewcommand{\shortauthors}{Lison and Kennington}

\begin{abstract}
Software architectures for conversational robots typically consist of multiple modules, each designed for a particular processing task or functionality. Some of these modules are developed for the purpose of making decisions about the next action that the robot ought to perform in the current context. Those actions may relate to physical movements, such as driving forward or grasping an object, but may also correspond to communicative acts, such as asking a question to the human user.  In this position paper, we reflect on the organization of those decision modules in human-robot interaction platforms. We discuss the relative benefits and limitations of modular vs. end-to-end architectures, and argue that, despite the increasing popularity of end-to-end approaches, modular architectures remain preferable when developing conversational robots designed to execute complex tasks in collaboration with human users. We also show that most practical HRI architectures tend to be either robot-centric or dialogue-centric, depending on where developers wish to place the ``command center'' of their system. While those design choices may be justified in some application domains, they also limit the robot's ability to flexibly interleave physical movements and conversational behaviours. We contend that architectures placing ``action managers'' and ``interaction managers'' on an equal footing may provide the best path forward for future human-robot interaction systems. 

\end{abstract}







\newcommand{\sds}[0]{\textsc{sds}}
\newcommand{\nlu}[0]{\textsc{nlu}}
\newcommand{\nlg}[0]{\textsc{nlg}}
\newcommand{\asr}[0]{\textsc{asr}}
\newcommand{\dm}[0]{\textsc{dm}}
\newcommand{\tts}[0]{\textsc{tts}}

\maketitle

\section{Introduction}

Conversational interaction allows humans to communicate with robots using a medium that is both intuitive and highly expressive (i.e., speech), but endowing robots with the ability to comprehend, represent, and produce spoken responses and integrate a speech system into the robots' own states and plans has been a subject of research for many years with limited success. The mere integration of an off-the-shelf automatic speech recognizer into the robotic platform is far from sufficient, as system developers need to address a host of additional challenges in communication, including how to handle turn-taking, dealing with noise, uncertainties and misunderstandings, and building common ground \cite{marge2022spoken}. Those phenomena, which extend well beyond speech processing, are precisely what the spoken dialogue systems (\sds) community has been addressing for many years. If humans are to naturally interact with robots, then robots need \sds --particularly, multimodal \sds, given the intrinsically multimodal nature of face-to-face interactions \cite{celiktutan2017multimodal}. 

Both robotic platforms and multimodal \sds\ are complex, multifaceted technical systems, but their integration in a single architecture raise a number of additional challenges that the community needs to face as robots that converse with humans using speech are expected to become more commonplace. In this position paper, we focus on one central question for autonomous systems: {\it decision making}. The problem of selecting the next action to take given the current state (as perceived by the robot sensors) has been widely studied in robotics \cite{schmerling2018multimodal} and has led to a wide range of specialized decision systems, from motion and path planning \cite{gasparetto2015path,mohanan2018survey} to object manipulation \cite{billard2019trends} and task planning \cite{hanheide2017robot}. Those decisions  may either rely on reactive mechanisms \cite{de2008behavior}, planning algorithms \cite{karur2021survey} or, increasingly, data-driven strategies optimised via imitation or reinforcement learning \cite{fang2019survey,singh2022reinforcement}.

\sds\ also need to make decisions about what to say next, given the current state of the conversation (as perceived/understood by the agent). Those decisions are typically made by a dedicated module called the {\it dialogue manager} \cite{young2013pomdp,Lison2015-ye}, sometimes also called the {\it interaction manager} in human--robot interaction \cite{abbasi2019multimodal}. The dialogue manager is responsible for deciding what to say and when to say it, and is itself typically divided in two sub-tasks: dialogue state tracking and action selection.  While the goal of dialogue state tracking is to maintain a representation of the current dialogue state given the stream of speech inputs and other observations, action selection relies on this state representation to select the most appropriate system response. 

A key question for action selection in both robotics and in multimodal \sds\ is how to operate in large state-action spaces. This question becomes particularly acute when both types of systems are combined into a single platform. In this position paper, we review some of the design choices related to the ``chain of command'' underpinning software architectures for conversational robots. In particular, we discuss the relative merits and limitations of modular architectures in light of the growing popularity of end-to-end approaches. We also observe that many HRI systems tend to adopt either a robot-centric or dialogue-centric view of the decision chain, often influenced by the particular research background of their developers. We argue that such design biases, while justifiable in some application domains, may also limit the system ability to handle complex situations including both physical and conversational tasks. The view advocated in this paper is that task/motion managers (responsible for the physical actions executed by the robot) and interaction managers (responsible for the robot's conversational behaviour) should ideally be placed side-by-side as part of a larger, modular architecture.

\section{Background}

In this section, we review some seminal and recent research literature on decision making in \sds\ and robotics. The reviews are by no means comprehensive, but should give the reader enough background to understand the architectural issues discussed in Section \ref{sec:discussion}.

\subsection{Decision-Making in SDS}

The \emph{Information State} approach to dialogue management \cite{larsson2000information,Traum2003-zl} has been highly influential in \sds\ research and continues to inspire modern dialogue systems. The approach revolves around a symbolic representation of the current dialogue state, which includes all types of information relevant for managing the system's conversational behaviour, such as questions under discussion, commitments, beliefs, and intentions from dialogue participants. Two sets of rules operate on this dialogue state: \begin{itemize}
    \item Update rules modify/extend the dialogue state on the basis of observations, such as new user inputs. 
\item Decision rules then select the most appropriate dialogue move(s) on the basis of the dialogue state.
\end{itemize}

\citet{Traum2003-zl} already note that there is no universal agreement on what should constitute a decision maker because different systems might carve up system modularity in different ways, some centralizing all decision-making in a single module while others rely on a more de-centralized distribution of labor (for example, to a natural language generation module that takes high-level information from the decision maker and generates an actual utterance). 

Early implementations of the Information State framework used symbolic representations of the dialogue state (and task/domain knowledge) along with handcrafted rules for both state update and action selection. Those handcrafted rules are now often replaced by data-driven models, as detailed in the large literature on dialogue state tracking  and dialogue policy optimization. Dialogue state tracking \cite{williams2016dialog,gao2019dialog,jacqmin2022you} seeks to predict the current dialogue state (typically represented as slot-value pairs) based on the stream of user inputs. Based on the current state, the next action is then selected by a dialogue policy optimized through supervised or reinforcement learning \cite{su2018reward,peng2018deep,zhang2022efficient}. 

Also seminal is the early work of \citep{Cohen1990-dk} on plan-based approaches to dialogue management, building on earlier work by \citep{Allen1979-tj}. Planning, the authors argue, is necessary if an agent is to be \emph{intentional}, defined as having choice with commitment.  Although plan-based frameworks are still used in a number of application domains where the task completion necessitates long-term reasoning \cite{santos2022review}, most current approaches rely on dialogue policies specifying a (learned) direct mapping between states and actions\footnote{This direct mapping from state to actions can be seen as representing precomputed plans for every possible state. This ``precomputation'' is often achieved through reinforcement learning, i.e.~by traversing large numbers of possible dialogue trajectories and using the resulting outcomes to estimate the expected cumulative reward $Q(s,a)$ of an action $a$ in a given state $s$. Those $Q$-values are now typically expressed as a neural network. By contrast, online planning algorithms determine the desirability of each action at runtime, using e.g.~forward planning \cite{ross2008online}.}. If the dialogue state is represented as probability distribution over possible slot-value pairs, this mapping can be formalized as a Partially Observable Markov Decision Process (POMDP), as shown by \citep{Young2010-bq2,Young2013-ny}. POMDPs have also been employed in robotics \cite{Whitney2017-sa}, although they are -- as in \sds\ research -- increasingly replaced by neural methods. 

End-to-end approaches that directly produce system responses from the user inputs (or their transcription in the case of speech-enabled interfaces) using encoder-decoder neural models are also increasingly popular \cite{zhang2018context,ghazvininejad2018knowledge,gao2019neural,zhang2020dialogpt}. Those approaches do not compute any explicit representation of the dialogue state, but encode the user inputs and the dialogue history into latent vector representations that are then employed to generate the system response token-by-token. Those end-to-end models are typically trained from dialogue corpora, either directly or on the basis of existing large neural language models. 

%We find it interesting to note that this very early work on \sds\ decisions included planning, and though planning is still used, much of the \sds\ field has split off into information state approaches, whereas in robotics planning continues to take center stage. 

\subsection{Decision-Making in Robotics}

Decision-making and planning constitute a central problem in robotics, encompassing various tasks from high-level task planning \cite{hanheide2017robot,karpas2020automated} to path planning \cite{gasparetto2015path,Hentout2022-pa} and to object manipulation/grasping and motion planning \cite{mohanan2018survey,billard2019trends,marturi2019dynamic}. Given a particular robot morphology and degrees of freedom, the planning algorithm seek to derive a path to an expected target configuration or location of the robot. This path is then employed to determine which actuators or motors to activate, and when. Robot planning can either rely on model-based approaches with formal planning languages such as PPDL and its variants \cite{fox2003pddl2,cashmore2015rosplan,canal2019probabilistic}, or employ data-driven methods, often based on (deep) reinforcement learning \cite{lei2018dynamic,aradi2020survey}. 

\cite{Fragapane2021-ko} recently reviewed the literature of planning and control of mobile robots that can navigate a space autonomously using one or more of various sensors (e.g., laser or vision-based guidance). Interestingly, in recent years autonomous mobile robots are increasingly relying on decentralized decision-making architectures (particularly for navigation). Planning is still central, but smaller decisions can be made by other modules more specific to a particular aspect of the robot, a claw for example. This is similar to work done in \sds, where many systems only use the decision making module for high-level decisions, though the issue of division of labor is by no means settled. 

Similar to, though distinct from, robotics is autonomous vehicle control, which also requires (fast, high-stakes) real-time decision making in novel situations. \cite{Veres2011-oj}'s review of the autonomous vehicle control literature identified a need to fill in a gap between integrating continuous sensor information and making decisions based on more abstract logic-based reasoning. Interestingly, they mention beliefs, goals, and intentions which have been important in planning and decision making literature in several fields, including \sds\ \cite{Cohen1990-dk}, as explained above. 

\subsection{Decision-Making in Human--Robot Interaction}
\label{sec:hri}

% does this go in the intro?

% A recent literature review (under review) surveyed different approaches to \sds\ decision making for human-robot interaction including hand-crafted, rule-based approaches and probabilistic approaches, as well as evaluation strategies. 

As a general trend, the choice of decision-making framework for human--interaction frameworks seems to be strongly influenced by the particular background and research interests of their developers. In other words, \sds\ researchers will often view the interaction manager as constituting the central decision module of their platform \cite{lison-kennington-2016-opendial,kennington2020rrsds,kiefer2020implementing}. Under this view, task planning is either subordinated to this interaction manager or encompassed into it. Similarly, robotics researchers will typically place task/motion planners at the center of their architecture and will often relegate dialogue management to a second-order component limited to handling speech inputs/outputs
\cite{cruz2018modular,fischer2018icub}. 

Much of the literature that reports on the integration of (multimodal) \sds\ in robotic platforms seems to be focused on either \sds\ or robotics, with few examples of practical systems integrating conversational abilities to robots executing complex physical tasks. Conversational robots are typically developed and evaluated in the context of social tasks that either do not require much movement, as in \cite{Kennington2014-if}) or where the movements are subordinated to communicative functions, such as gestures \cite{kopp2008multimodal,wagner2014gesture} or facial expressions \cite{Al_Moubayed2012-rj}.

%Another recent review by \cite{Tellex2020-tl} looked at robots that use language with focus on language grounding---a critical part of spoken interaction between robots and humans---though not on decision making (focus is put on the task of navigation using a linear temporal logic). 

\section{Open questions}
\label{sec:discussion}

In this section, we review two key questions related to the design of software architectures for conversational robots: \begin{itemize}
\item where to adopt modular vs. end-to-end architectures ;
\item how the decision modules should interact with one another.
\end{itemize}

\subsection{Modular vs. End-to-end SDS}

End-to-end \sds\ rely on a single model -- typically a neural network -- that takes an (often text-based) input and directly produce an output, such as a question-answering system that provides an answer to a question, or a social chatbot that produces responses given text input. End-to-end systems often focus on the ability to produce a plausible response to any input, based on an encoder-decoder model trained or fine-tuned from dialogue data. The ``decisions'' made by the model correspond in this case to the successive selection of tokens in the response. In other words, end-to-end conversational models do not have any explicit concept of task or plan -- they are only optimized to produce plausible responses. As such, their use for task-oriented systems remains challenging, although see \cite{wen2017network,yang2021ubar} for a few attempts. 

In contrast, modular systems are made up of modules that have well-defined roles in the system, and which can be made to communicate with each other. For example, a prototypical \sds\ is often made up of a set of modules including automatic speech recognition (\asr) that transcribes speech to a text representation of the human utterances, natural language understanding (\nlu) that takes the text and yields a semantic abstraction, dialogue management (including dialogue state tracking) that makes the high-level decision about the next action to take given the semantic abstraction, natural language generation (\nlg) that takes the dialogue manager's decision and determines which words to use and in what order, and text-to-speech (\tts) which actually utters the response.

Both types of approaches have their advantages and limitations. End-to-end systems are conceptually simpler (even though their neural architectures may be large and composed of many layers), and do not suffer from cascading failures as in modular systems. As they are fully differentiable, they may also be optimized directly from the observed inputs and outputs, without needing to rely on annotations or semantic abstractions. 

However, end-to-end approaches also have a number of shortcomings. Most importantly, from a designer perspective, their conversational behaviour is much more difficult to control, adjust, and interpret. System designers can never be sure of what the system will end up producing as response, and the tendency of generative language models to e.g.~hallucinate facts is well documented \cite{ji2022survey}. As they are primarily text-based, their integration with non-linguistic knowledge sources may also be challenging, although there exists a number of approaches integrating neural models with knowledge bases \cite{ghazvininejad2018knowledge,gao2019dialog} or image/videos \cite{le2019multimodal,sundar-heck-2022-multimodal}. Finally, even though few-shots learning approaches may be employed to mitigate problems of data scarcity \cite{10.1145/3386252,peng2020few}, their portability to scenarios with no or limited data remains difficult. 

In our view, end-to-end models may be useful for speech-enabled robots whose primary purpose is to serve as social companion. However, modular approaches remain preferable for conversational robots set to execute physical tasks in the real world, either alone or in collaboration with humans. This position is motivated by several considerations. Most importantly, task-oriented robots are themselves developed around modular architectures such as ROS \cite{Quigley2009}, and a modular \sds\ is much easier to integrate in those platforms than a monolithic, end-to-end conversational model. This is particularly true when the interaction management needs to be tightly interfaced with other decision-making modules for task, path or motion planning, which need to rely on dedicated representations and algorithms and cannot be reduced to a token-by-token generation task. In addition modular approaches also provide better control and interpretability, which is a decisive factor when developing complex robotic platforms operating in the real world. 

\begin{figure*}
  \center
  \includegraphics[width=0.95\textwidth]{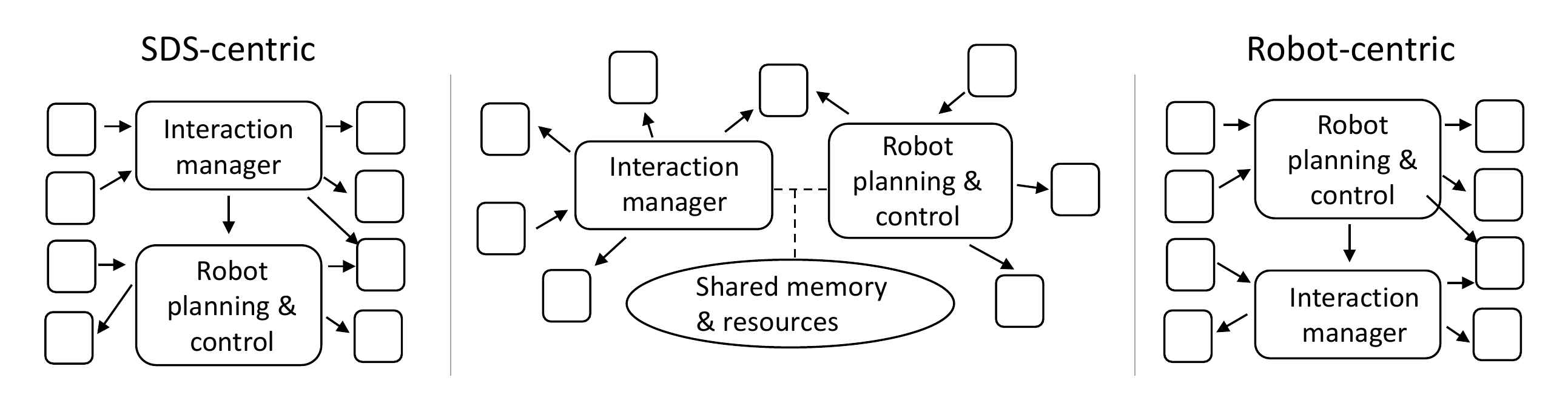}
  \caption{Architectural choices for decision-making components in conversational human--robot interaction platforms. The left side shows a \sds-centric view in which the interaction manager is the top decision-making module, and the right side a robot-centric view where the robot planners and controllers take precedence. At the center, a schematic architecture in which both decision-making modules are set on an equal footing and operate on the basis of a shared memory and common resources.}
  \label{fig:dmdims}
\end{figure*}

\subsection{Interactions between Decision-Making Modules}

Which modules should be responsible for making decisions about the (linguistic and non-linguistic) actions that can be performed by the robot? As noted in Section \ref{sec:hri}, most current approaches tend to adopt either a \sds-centric view or a robot-centric view of the decision chain. In the \sds-centric view, the main center of command is the interaction manager, which is typically retooled to include decisions for robot actions. In contrast, a robot-centric view places the task/motion planning as the center stage, only including the \sds\ as a secondary subsystem limited to receiving speech inputs and/or communicating information back to the user. 

Both design choices may certainly be appropriate in certain application domains. For instance, \sds-centric architectures may work well for social robots designed to provide stimulating social interactions with their human interlocutors. In those cases, managing the conversational behaviour of the robot is indeed the main objective to which all modules are subordinated. Similarly, robot-centric architectures are well-suited for robots that are primarily developed for the execution of a particular physical task, and only use \sds\ to convey to information to its human users. 

However, we contend that such architectures are insufficient for robotic systems that aim to achieve complex, collaborative tasks including mixed-initiative dialogues with human users. Indeed, such scenarios will often interleave task execution and spoken interactions in complex, often unpredictable ways. This interplay of behaviours is difficult to address in a single decision-making module, and is better handled through an architecture where the interaction manager operates in parallel to the robot planners and action controllers. Conceptual visualizations for some of these different decision making architectures can be seen in Figure~\ref{fig:dmdims}.

The interaction manager and the other robot-related components need, however, to interact closely, as their actions may impact one another. For instance, the robot may not be able to make gestures while in motion. In addition, the interaction manager may need to obtain information about the physical state of the robot and/or status of the task to complete in order to answer questions uttered by the user. As shown in Figure~\ref{fig:dmdims}, one way to allow for an efficient exchange of information is to rely on a shared memory that can be read/written by those decision-making modules, and can also be used to resolve potential conflicts between the actions selected by each component. 

Recent work on the EVA platform shows an impressive system that is equal parts \sds- and ``robot-" (i.e., virtual agent-) centric in that both the decisions made about system speech acts and virtual agent articulation are both treated as first-class citizens and important to the unfolding interaction.\footnote{\url{https://openstream.ai/}} The underlying model is plan-based and, from our understanding, despite being fairly modular is quite centralized in how decisions are made. Modeling such a system of course is challenging as state and actions spaces can potentially be prohibitively large. Other work has used a centralized module for dialogue and robot action decision making, though the set of actions were fairly limited \cite{Lison2012-nx,Torres-Foncesca2022-ry}

\section{Discussion \& Conclusion}

% do we have an opinion on which one we prefer?

Decision making in \sds\ and robotics is a very important part of a workable system in both fields. In this paper, we looked at how the fields differ, and offered some possible paths forward on how to model decision making when \sds\ and robots are brought together. In order for the community to share resources such as models and platforms, it is prudent to find a principled solution to what often seems to be ad-hoc modeling decisions. We have by no means comprehensively solved the problem of determining how decision-making components should interact in complex human--robot interaction systems. Moreover, we did not discuss how different state representations (i.e., how inputs and actions are represented in a system) can affect modeling decisions, nor did we touch on the importance of time horizons in decision making (i.e., some decisions are near-term, while others are longer-term like in chess where there is always a next move, but there is also an overall strategy that includes planning moves beyond just the next one). There may be a way to systematically test different methods of decision making to determine important trade offs using a task that is equal parts dialogue and robot control, such as \cite{Padmakumar2021-jx}, a virtual robot recipe task, which has been annotated with dialogue acts \cite{Gella2022-tb}, which we leave for future work.

\bibliographystyle{ACM-Reference-Format}
\bibliography{paperpile, refs, software}

% \appendix

\end{document}